\PassOptionsToPackage{table}{xcolor}

\documentclass{article}

\usepackage{microtype}
\usepackage{graphicx}
\usepackage{subcaption}
\usepackage{booktabs} 

\usepackage{hyperref}

\usepackage[preprint]{icml2026}

\usepackage{amsmath}

\usepackage{amssymb}
\usepackage{mathtools}
\usepackage{amsthm}

\usepackage{siunitx}
\usepackage[table]{xcolor}
\usepackage{multirow}
\usepackage{pifont}
\definecolor{rowblue}{RGB}{230,245,255}
\newcommand{\cmark}{\ding{51}}
\newcommand{\xmark}{\ding{55}}

\usepackage[capitalize,noabbrev]{cleveref}

\theoremstyle{plain}

\theoremstyle{definition}

\theoremstyle{remark}

\usepackage[disable,textsize=tiny]{todonotes}

\icmltitlerunning{TrackRef3D: Multi-View Consistent Track-then-Label for Open-World Referring Segmentation in 3D Gaussian Splatting}

\begin{document}

\twocolumn[
  \icmltitle{TrackRef3D: Multi-View Consistent Track-then-Label for Open-World Referring Segmentation in 3D Gaussian Splatting}



  \icmlsetsymbol{equal}{*}

  \begin{icmlauthorlist}
    \icmlauthor{Yuyang Tan}{ecnu}
    \icmlauthor{Renhe Zhang}{ecnu}
    \icmlauthor{Hang Zhang}{shai}
    \icmlauthor{Ao Li}{uestc}
    \icmlauthor{Xin Tan}{ecnu,shai}
  \end{icmlauthorlist}

  \icmlaffiliation{ecnu}{East China Normal University, Shanghai, China}
  \icmlaffiliation{shai}{Shanghai AI Laboratory}
  \icmlaffiliation{uestc}{University of Electronic Science and Technology of China, Chengdu, China}

  \icmlcorrespondingauthor{Xin Tan}{xtan@cs.ecnu.edu.cn}

  \icmlkeywords{3D Gaussian Splatting, Referring Segmentation, Multi-View Consistency, Open World}
  \vskip 0.3in
]



\printAffiliationsAndNotice{}  

\begin{abstract}
 Referring 3D Gaussian Splatting (R3DGS), which utilizes natural language for 3D object segmentation, has emerged as a crucial capability for embodied AI. However, existing methods typically rely on expensive per-scene manual annotation and per-view pseudo mask generation, which suffer from multi-view inconsistency and poor generalization to varying query specificities. To address this, we present TrackRef3D, a fully automatic pipeline that achieves open-world referring segmentation in 3D Gaussian Splatting (3DGS) without manual annotation by introducing a multi-view consistent track-then-label paradigm that fundamentally decouples object discovery from semantic grounding. Specifically, we propose a Trajectory-Aware Semantic Consensus Module (TSCM) which aggregates cross-view predictions via synonymous clustering and trajectory-aware voting to establish a canonical semantic identity, thereby ensuring multi-view consistency. Furthermore, we employ a visibility-aware description generation strategy to mitigate ambiguity and propose a Hybrid Training Strategy (HTS) that jointly optimizes coarse category semantics and fine-grained referential cues to ensure robustness under varying query specificities using a multi-positive contrastive objective. Extensive experiments on benchmarks demonstrate that TrackRef3D achieves state-of-the-art performance.
\end{abstract}

\section{Introduction}

\begin{figure}[ht]
    \centering\centerline{\includegraphics[width=\columnwidth]{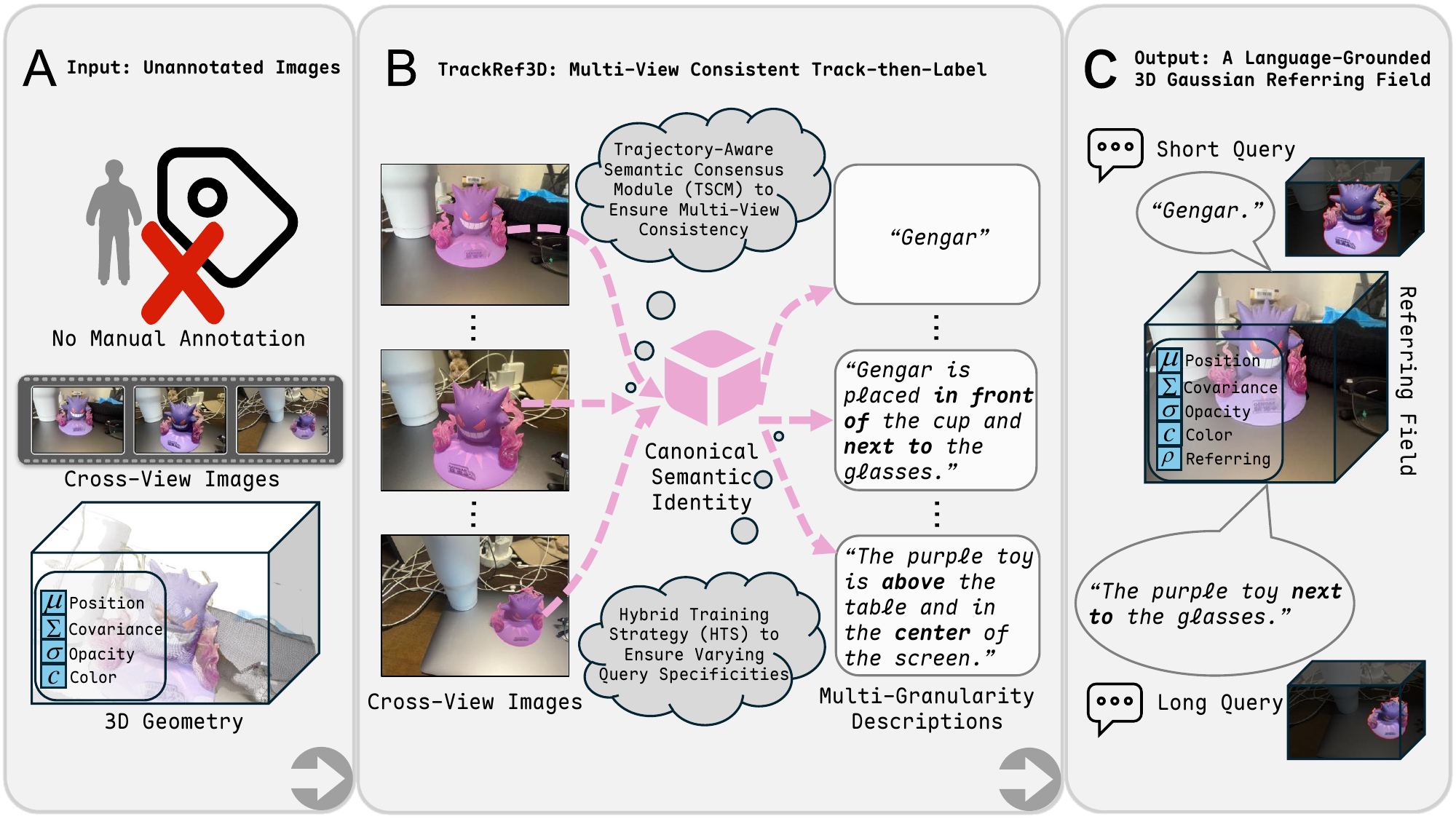}}
    \caption{Given only unannotated images and a reconstructed 3D Gaussian field for geometry, we perform the Trajectory-Aware Semantic Consensus Module (TSCM) to obtain canonical semantic identities and multi-granularity descriptions supervised by the Hybrid Training Strategy (HTS). The result is a language-grounded 3D Gaussian referring field that supports short and long queries.}
    \label{fig:overview}
\end{figure}

\begin{figure}[ht]
\centering
\centerline{\includegraphics[width=\columnwidth]{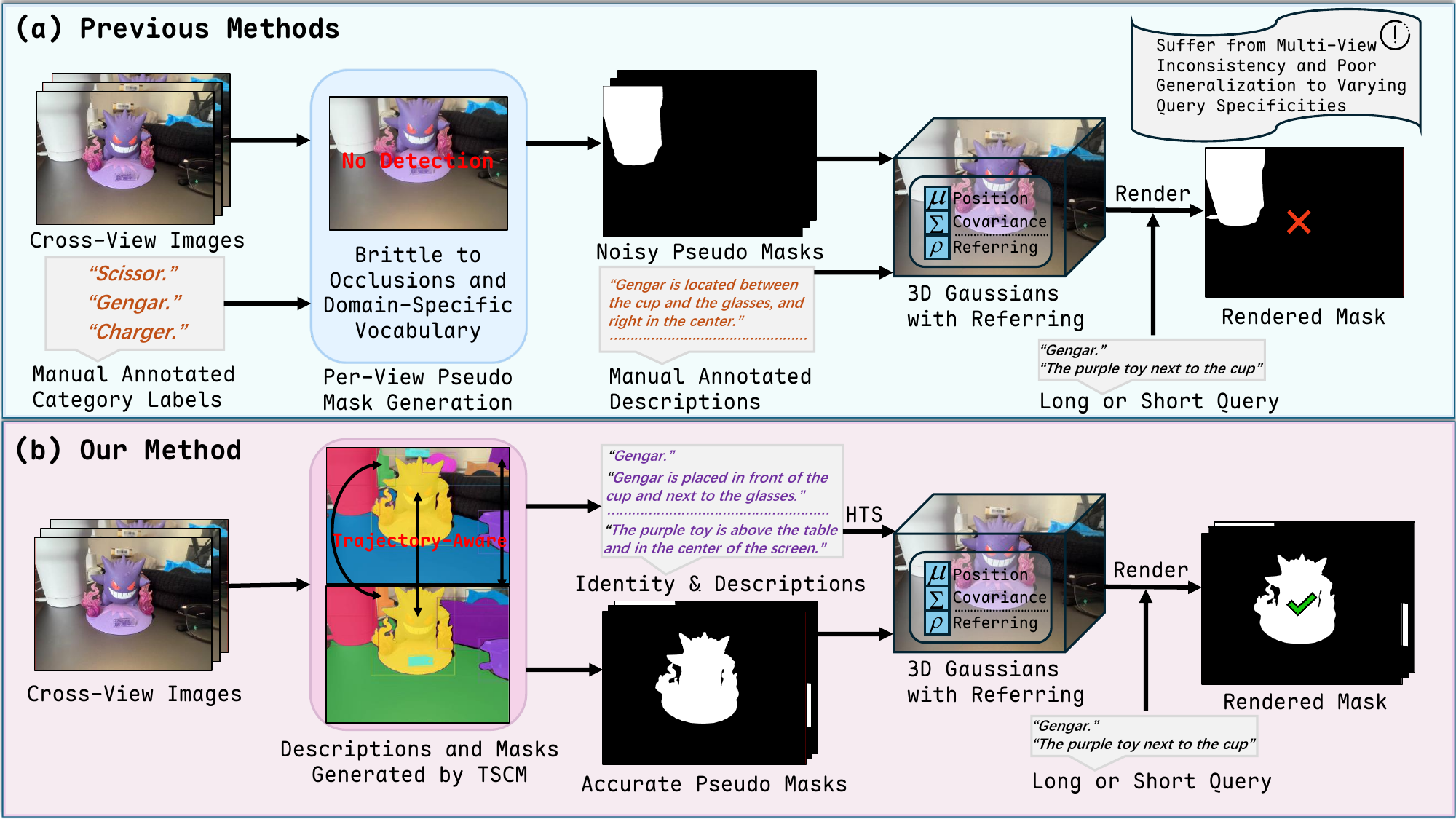}}
\caption{(a) Previous methods rely on manually annotated category labels for per-view pseudo mask generation and referring descriptions, which are brittle to occlusions and domain-specific vocabulary, leading to multi-view inconsistent supervision. (b) TrackRef3D automatically generates cross-view consistent pseudo masks and multi-granularity descriptions via video tracking with clustering and voting, enabling consistent supervision for training a language-grounded 3D Gaussian referring field.}
\label{fig:motivation}
\end{figure}

Referring 3D Gaussian Splatting (R3DGS) \cite{ReferSplat}, which utilizes natural language to segment objects that may be occluded or invisible from novel viewpoints, has emerged as a crucial capability for embodied AI. Such language-conditioned 3D referring segmentation is broadly useful for robots \cite{Werby-RSS-24,zhu2025mtu} and autonomous driving systems \cite{shao2023lmdrive,DriveVLM} to perceive and interact with environments through intuitive linguistic commands. In the existing R3DGS pipeline, constructing a language-grounded 3D Gaussian referring field typically relies on manually annotated labels and scene-level referring descriptions, which are used to perform per-view detection and segmentation for pseudo mask supervision, and then to train the referring field to align 3D Gaussians with the provided descriptions.

However, following this manual annotation pipeline, recent R3DGS methods such as ReferSplat \cite{ReferSplat}, as shown in Fig.~\ref{fig:motivation} still face three fundamental obstacles when scaling to real-world scenarios. First, they require human-annotated category labels and referring descriptions for each scene, which is prohibitively expensive and makes per-scene annotation unrealistic in real-world deployments. Second, their supervision is generated in a per-view manner: 2D foundation models \cite{ren2024groundingdino, ren2024groundedsam, ravi2024sam2} may miss the target under occlusions and domain-specific vocabulary, then generate inconsistent pseudo masks across views, leading to contradictory supervision known as multi-view inconsistency. Third, training dominated by referring descriptions biases the model toward long descriptions and degrades generalization to short prompts that users naturally query in practice. These issues collectively motivate an open-world pipeline that (i) eliminates manual per-scene annotation, (ii) enforces cross-view semantic consensus by aggregating evidence along the camera trajectory, and (iii) synthesizes multi-granularity supervision signals to ensure robust generalization across varying query specificities.

We present TrackRef3D (Fig.~\ref{fig:overview}), a fully automatic pipeline that achieves open-world referring segmentation in 3DGS without manual annotation by introducing a multi-view consistent track-then-label paradigm that fundamentally decouples object discovery from semantic grounding. In the object discovery phase, we first leverage 2D foundation models to obtain potential object masks and labels, then employ video tracking to associate these masks into coherent trajectories across the full camera path. This process yields stable track IDs despite occlusions and domain-specific vocabulary, serving as the robust foundation for our subsequent semantic grounding. With stable trajectories established, we propose the Trajectory-Aware Semantic Consensus Module (TSCM) to construct cross-view consistent supervision. For each trajectory, we collect multi-view semantic predictions and perform synonymous clustering to merge semantically equivalent labels into the same group, followed by the trajectory-aware voting to select a canonical semantic identity that is propagated to all views of the track. In parallel, we exploit visibility-aware description generation with vision-language models to generate referring descriptions based on the keyframe where the object is most visible, reducing ambiguity from occluded views and ensuring that the generated descriptions remain consistent across viewpoints. Finally, to ensure robust open-world deployment, we propose the Hybrid Training Strategy (HTS) that addresses overfitting to long descriptions and ensures robustness under varying query specificity. This strategy jointly optimizes short category semantics and multi-granularity descriptions for each track using a multi-positive contrastive objective, treating all descriptions associated with the same trajectory as positive pairs so the model learns both coarse category semantics and fine-grained referential cues. We conduct extensive experiments on benchmarks and achieve state-of-the-art performance. 

In summary, our main contributions are:

\begin{itemize}
  \item We introduce a fully automatic multi-view consistent track-then-label paradigm that achieves open-world referring segmentation in 3D Gaussian Splatting.
  \item To address multi-view inconsistency, we propose a Trajectory-Aware Semantic Consensus Module which aggregates cross-view predictions via clustering and voting to establish a canonical semantic identity to ensure consistent supervision. 
  \item To ensure robust open-world deployment under varying query specificity, we propose a Hybrid Training Strategy that jointly optimizes coarse category semantics and fine-grained referential cues using a multi-positive contrastive objective, effectively preventing the issue of overfitting to long descriptions.
  \item We conduct extensive experiments on Ref-LERF, LERF-OVS and 3D-OVS benchmarks and a self-collected indoor Laboratory scene. Comprehensive experiments show that TrackRef3D achieves the best performance against previous methods.
\end{itemize}

\section{Related Work}

\subsection{Open-Vocabulary 3D Gaussian Semantic Field}

3DGS \cite{kerbl3Dgaussians} introduces an explicit scene representation with real-time rendering, which has recently been extended with semantic features for open-vocabulary 3D scene understanding \cite{shi2023language,zuo2024fmgs,Lowis3D,drsplat25,CAGS,chiou2026profuseefficientcrossviewcontext}. A common paradigm is to lift 2D vision-language features (often extracted with CLIP \cite{radford2021learning}) into 3D Gaussians and answer a short query by matching rendered features with the text embedding \cite{qin2024langsplat,zhou2024feature,liang2024supergsegopenvocabulary3dsegmentation,jsmbankILGS,Lu2025Segment,vala,OpenInsGaussian}. Within this line, methods differ mainly in how the 2D semantics are distilled across views and lifted to Gaussians, as well as what downstream capabilities are emphasized such as open-vocabulary segmentation and editing.

Beyond directly distilling per-view semantics, several works move toward more point-level open-vocabulary understanding by improving the 2D--3D association and leveraging instance cues. OpenGaussian \cite{wu2024opengaussian} targets point-level open-vocabulary understanding and highlights that previous approaches largely focus on pixel-level parsing, where the semantics can be affected by 2D--3D associations. Gaussian Grouping \cite{gaussian_grouping} augments each Gaussian with a compact identity encoding and supervises it with 2D mask predictions to enable open-world segmentation and editing. GOI \cite{qu2024goi} further proposes query-adaptive selection of relevant Gaussians by treating feature selection as an optimizable hyperplane separation problem in the semantic feature field. 

Overall, while these methods substantially advance open-vocabulary understanding in 3DGS, they do not explicitly target referring segmentation that requires grounding referring descriptions to a specific instance across viewpoints, motivating referring segmentation in 3DGS as a distinct setting \cite{ReferSplat}.

\subsection{Language-Grounded 3D Referring Segmentation}

Referring segmentation in 3D point clouds (3DRES) aims to segment the specific target instance described by a natural language description \cite{3D-STMN,RG-SAN,wu20243dgresgeneralized3dreferring,liu2024less,IPDN,mvggt}. Early representative efforts develop this setting with relational reasoning on object candidates, such as text-guided graph reasoning in TGNN \cite{TGNN} and structured cross-modal graph modeling in X-RefSeg3D \cite{X-RefSeg3D}. More recent approaches further strengthen multi-modal fusion with geometry-aware interactions \cite{RefMask3D}.

To bring referring segmentation into 3DGS with real-time rendering, ReferSplat \cite{ReferSplat} introduces R3DGS, which learns a language-grounded 3D Gaussian referring field to render referring segmentation masks for targets that may be occluded or invisible from novel viewpoints. Nevertheless, such a pipeline typically requires expensive per-scene manual annotation and per-view pseudo mask generation, which can lead to multi-view inconsistency when the same object receives contradictory pseudo masks across viewpoints. These limitations motivate an open-world R3DGS pipeline that avoids per-scene manual annotation and enforces cross-view consistent supervision.

\section{Method}
\begin{figure*}[t]
  \centering
  \includegraphics[width=0.95\linewidth]{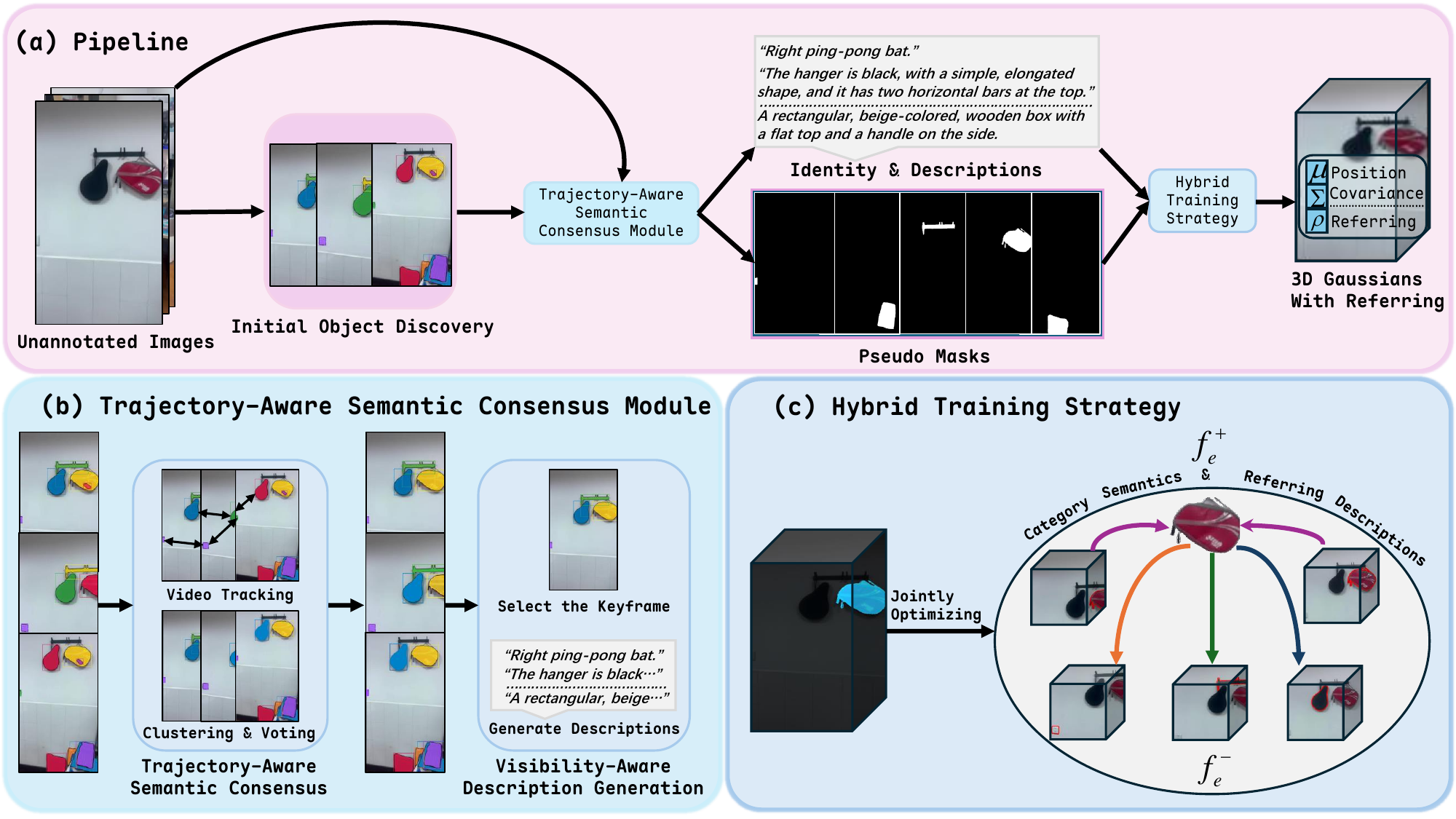}\vspace{0mm}
  \caption{Overview of TrackRef3D. Starting from unannotated cross-view images, we first generate initial masks and labels. The Trajectory-Aware Semantic Consensus Module (TSCM) associates per-view masks into stable trajectories via video tracking and ensures multi-view consistency by establishing a canonical semantic identity through clustering and voting. It further generates descriptions based on the selected keyframe. Finally, we jointly optimize a language-grounded 3D Gaussian referring field using the Hybrid Training Strategy (HTS), which aligns Gaussians with both the coarse category semantics and fine-grained referential cues to ensure robustness under varying query specificities.}
  \label{fig:pipeline}
  \vspace{0mm}
\end{figure*}

\subsection{Overview}
\textbf{Training.} Given a set of RGB images $\{I_i\}_{i=1}^{N}$ and a reconstructed 3D Gaussian scene for geometry $\mathcal{G}$, TrackRef3D trains a language-grounded 3D Gaussian referring field $\mathcal{G}_{\text{ref}}$. Specifically, we refine these raw signals into cross-view consistent supervision via the TSCM. The high-quality supervision then guides the training of a language-grounded 3D Gaussian referring field $\mathcal{G}_{\text{ref}}$ through the HTS, ensuring robustness under varying query specificity.

\textbf{Inference.} Given an arbitrary description and camera parameters specifying the target viewpoint, TrackRef3D renders the corresponding 2D segmentation mask by computing word-to-Gaussian similarity in the learned language-grounded 3D Gaussian referring field $\mathcal{G}_{\text{ref}}$.

As illustrated in Fig.~\ref{fig:pipeline}, our pipeline consists of three stages. 
(1) \textbf{Initial Object Discovery} (Sec.~\ref{sec:3.2}) discovers potential object labels and masks.
(2) \textbf{Trajectory-Aware Semantic Consensus Module} (Sec.~\ref{sec:3.3}) ensures multi-view consistency by establishing a canonical semantic identity and generates multi-granularity descriptions for supervision.
(3) \textbf{Hybrid Training Strategy} (Sec.~\ref{sec:3.4}) ensures robustness under varying query specificities by jointly optimizing short category semantics and multi-granularity descriptions using a multi-positive contrastive loss.

\subsection{Initial Object Discovery}
\label{sec:3.2}
The first stage of our pipeline aims to discover all potential objects in the scene from individual views. We leverage foundation models \cite{xiao2023florence,ravi2024sam2} that combine object detection and segmentation capabilities to extract potential object masks and labels. Specifically, we first apply open-vocabulary detection to identify object bounding boxes along with their predicted category labels in each view. These bounding boxes are then fed as prompts to a segmentation model to generate precise pixel-level masks for each detected object.

This per-view processing enables us to discover a wide range of objects in the scene at the instance-level without requiring predefined categories. Crucially, the open-vocabulary allows us to detect objects with domain-specific vocabulary that previous detectors would miss. However, this approach inevitably introduces two types of limitations: (1) \textbf{noisy pseudo masks} due to occlusions and (2) \textbf{inconsistent category labels} for the same object across different views. For instance, a bowl of ramen may be labeled as ``ramen'' in one view but ``bowl'' or ``food'' in another due to cross-view predictions. These inconsistencies stem from the fundamental limitation of per-view processing: without cross-view video tracking, 2D foundation models cannot maintain semantic consistency for objects undergoing appearance changes across the camera trajectory.

Rather than attempting to filter out these inconsistencies at this stage, we preserve all detected predictions, deferring the resolution of cross-view semantic inconsistency to the TSCM in the next stage. This design allows us to maximize object recall while ensuring consistent supervision.

\subsection{Trajectory-Aware Semantic Consensus Module}
\label{sec:3.3}
We draw inspiration from the track-centric labeling paradigm \cite{spam2024eccv}, which shifts the annotation focus from individual views to coherent trajectories to capture long-term spatiotemporal dependencies. To resolve the multi-view inconsistency introduced by per-view processing, we introduce a trajectory-aware mechanism that operates at the camera trajectory rather than individual views. The key insight is that the same object should maintain a consistent semantic identity across its entire trajectory, regardless of viewpoint changes or occlusions. Our approach consists of four stages: video tracking, synonymous clustering, trajectory-aware voting, and description generation.



\textbf{Video Tracking.} We first employ video tracking to associate per-view masks into temporally coherent trajectories. The tracker establishes correspondences across views by matching visual features and spatial locations, assigning each mask instance a unique track ID. This produces a set of object trajectories $\mathcal{T} = \{T_1, T_2, ..., T_K\}$, where each trajectory $T_i$ contains all mask instances of the same object across views. Critically, the tracker leverages temporal context to maintain identity even when objects undergo occlusions and re-enter the viewpoint that per-view methods fundamentally lack, ensuring that semantic grounding is performed on coherent instances rather than isolated fragments.

\textbf{Synonymous Clustering.} Before applying the trajectory-aware voting, we address a subtle but important issue: linguistic variations in category names. The initial per-view predictions may label the same object concept with slight variations. We obtain text embeddings for all detected category names using a sentence embedding model \cite{sentencetransformer}, and measure semantic similarity with cosine similarity. We implement clustering using agglomerative clustering (average linkage) on the precomputed cosine-distance matrix $d(c_i,c_j)=1-\cos(e_i,e_j)$, with the distance threshold set to $1-\tau_{\text{sem}}$. For each cluster, we select the shortest surface form as the clustered identity.

\textbf{Trajectory-aware Voting.} For each trajectory $T_i$, we collect the clustered identities after synonymous clustering and count their occurrence frequencies across views. Specifically, we denote the clustered identity at view $t$ as $\hat{c}_t = \phi(c_t)$, where $\phi(\cdot)$ maps a raw per-view label to its clustered identity by grouping synonymous labels whose text-embedding similarity is above $\tau_{\text{sem}}$. The most frequent clustered identity is selected as the canonical semantic identity $c_i^*$ for the entire trajectory:
\begin{equation}
c_i^* = \arg\max_{c \in \mathcal{C}_i} \sum_{t \in T_i} \mathbb{I}[\hat{c}_t = c],
\end{equation}
where $\mathcal{C}_i=\{\hat{c}_t \mid t \in T_i\}$ is the set of clustered identities observed in trajectory $T_i$, and $\mathbb{I}[\cdot]$ is the indicator function. The summation iterates over all views in the trajectory to count occurrences of each clustered identity. The winning identity $c_i^*$ is then propagated to all views where the object appears, replacing all clustered identities. This ensures complete multi-view consistency where every mask belonging to the same trajectory receives cross-view consistent  supervision.

\textbf{Description Generation.} A key challenge is that naively selecting a random view for description generation can produce ambiguous or misleading descriptions when objects are heavily occluded. To address this, we introduce a visibility-aware sampling that conditions description generation on the keyframe where the object is most clearly visible.

For each trajectory $T_i$ containing masks across multiple views, we compute a visibility score that balances object visibility and spatial context. A critical observation is that neither extremely large masks nor extremely small masks provides optimal descriptive quality. Instead, views with moderate mask areas tend to capture the object with sufficient details while preserving contexts for spatial reasoning. We formalize this intuition through a visibility score that penalizes scale deviation from the median area:
\begin{equation}
v_t = A_t \cdot \exp\left(-\frac{(\sqrt{A_t} - \sqrt{A_{med}})^2}{2\sigma^2}\right),
\end{equation}
where $A_t$ is the mask area at view $t$, $A_{med}$ is the median area across all views in trajectory $T_i$, and $\sigma$ controls the tolerance for area deviation. The first term $A_t$ ensures sufficient visibility, while the exponential term $\exp(\cdot)$ applies a Gaussian-like weighting in $\sqrt{A_t}$ that attains its maximum when $A_t=A_{med}$, penalizing extreme scale deviations. Combined with the multiplicative factor $A_t$, the overall score trades off sufficient visibility and moderate scale to select a clear yet context-preserving view. We then select the keyframe with maximum score:
\begin{equation}
t_i^* = \arg\max_{t \in T_i} v_t.
\end{equation}
This scoring mechanism naturally selects the view where the object is neither too close nor too far, corresponding to viewpoints where both the object's attributes and its spatial relationships to surrounding objects are most clearly observable. This ensures that the vision-language model \cite{hong2024cogvlm2} receives optimal visual input for generating descriptions that accurately capture both object visibility and spatial context.

\subsection{Hybrid Training Strategy}
\label{sec:3.4}

The final stage addresses the issue where models trained exclusively on long descriptions fail to generalize to short queries. We posit that a robust referring field $\mathcal{G}_{\text{ref}}$ should activate the same 3D representation regardless of varying query specificities. Leveraging the canonical semantic identity $c_i^*$ established by the TSCM, we create a unified semantic anchor to align 3D Gaussians with both coarse category semantics and fine-grained referential cues.

We optimize the model using a multi-positive contrastive objective \cite{khosla2020supervised}. Unlike standard contrastive learning, we explicitly treat both the category semantics and referring descriptions of a track as positive pairs. For a given object with the canonical semantic identity $c_i^*$, let $f_g$ denote the positive Gaussian embedding corresponding to a text query, obtained by averaging the referring features of the Gaussians selected by the rendered mask, and let $f_e^+$ denote the text embeddings of all positive descriptions including the category semantics and its referring descriptions. The contrastive loss is formulated as:
\begin{equation}
\mathcal{L}_{\text{con}} = - \frac{1}{|\mathcal{P}(c_i^*)|} \sum_{f_e^{+} \in \mathcal{P}(c_i^*)} \log \frac{\exp(f_g \cdot f_e^{+} / \tau)}{\sum_{f_e \in \mathcal{D}} \exp(f_g \cdot f_e / \tau)},
\end{equation}

where $\mathcal{P}(c_i^*)$ denotes the set of positive descriptions specifically associated with identity $c_i^*$ across granularities, while $\mathcal{D}$ represents the union of all descriptions in the training batch, serving as the contrastive pool containing negatives from other identities, and $\tau$ is the temperature parameter. Critically, this formulation treats the category semantics and referring descriptions as equally valid positives, enforcing that the referring field learns both coarse category semantics and fine-grained referential cues. The total training loss $\mathcal{L}$ combines the contrastive loss $\mathcal{L}_{\text{con}}$ with a segmentation loss  $\mathcal{L}_{\text{seg}}$ that ensures spatial accuracy:
\begin{equation}
\mathcal{L} = \mathcal{L}_{\text{seg}} + \lambda \mathcal{L}_{\text{con}},
\end{equation}
where $\mathcal{L}_{\text{seg}}$ is a binary cross-entropy loss between the rendered masks and the pseudo masks produced by TSCM, and $\lambda$ controls the balance between pixel-level segmentation and semantic grounding. This design closes the loop of our pipeline: the TSCM ensures consistent object identity, while the HTS ensures consistent language grounding under varying query specificities.

\section{Experiments}
\subsection{Experimental Setting}

\textbf{Datasets.} We evaluate TrackRef3D on three established benchmarks and one self-collected Laboratory scene to demonstrate its effectiveness in open-world scenarios. (1) Ref-LERF \cite{ReferSplat} is a referring segmentation benchmark containing four real-world scenes with human-annotated referring descriptions. (2) LERF-OVS \cite{lerf2023,qin2024langsplat} and 3D-OVS \cite{liu2023weakly} are open-vocabulary benchmarks annotated with canonical category labels. (3) Laboratory is a self-collected indoor scene with 1033 views for training and 10 views for testing. For training, all supervision signals are obtained entirely using our automatic pipeline, serving as a direct validation of our method's open-world capability and generalization to an unseen environment. For evaluation, we provide human-annotated 2D masks on the 10 test views as ground truth for mIoU computation, and we use human-annotated descriptions for testing. These test descriptions are used only for evaluation and are never involved in training.

\begin{table}[t]
\centering
\caption{Comparison on the Ref-LERF dataset with state-of-the-art methods in terms of \textbf{mIoU ($\uparrow$)}.}
\label{tab:ref-lerf}
\setlength{\tabcolsep}{4.4pt} 
\renewcommand{\arraystretch}{1.00} 
\begin{tabular}{ l|
S[table-format=2.1,detect-weight=true,detect-inline-weight=math]
S[table-format=2.1,detect-weight=true,detect-inline-weight=math]
S[table-format=2.1,detect-weight=true,detect-inline-weight=math]
S[table-format=2.1,detect-weight=true,detect-inline-weight=math]
|
S[table-format=2.1,detect-weight=true,detect-inline-weight=math] }
\toprule
\textbf{Method} & \textbf{ram.} & \textbf{fig.} & \textbf{tea.} & \textbf{kit.} & \textbf{avg.} \\
\midrule
SPIn-NeRF         &  7.3 &  9.7 & 11.7 & 10.3 &  9.8 \\
LangSplat         & 12.0 & 17.9 &  7.6 & 17.9 & 13.9 \\
GS-Grouping       & 27.9 &  8.6 & 14.8 &  6.3 & 14.4 \\
Grounded SAM      & 14.1 & 16.0 & 16.9 & 16.2 & 15.8 \\
GOI               & 27.1 & 16.5 & 22.9 & 15.7 & 20.5 \\
ReferSplat        & 35.2 & 25.7 & 31.3 & 24.4 & 29.2 \\
\rowcolor{rowblue}
\textbf{TrackRef3D (Ours)} & \textbf{45.7} & \textbf{34.2} & \textbf{41.7} & \textbf{33.6} & \textbf{38.8} \\
\bottomrule
\end{tabular}
\end{table}

\begin{table}[t]
\centering
\caption{Comparison on the Laboratory scene with state-of-the-art methods in terms of \textbf{mIoU ($\uparrow$)}.}
\label{tab:laboratory}
\renewcommand{\arraystretch}{1.00}
\setlength{\tabcolsep}{9.5pt}
\begin{tabular}{l
S[table-format=2.1,detect-weight=true,detect-inline-weight=math]
S[table-format=2.1,detect-weight=true,detect-inline-weight=math]}
\toprule
\textbf{Method} & \textbf{Referential} & \textbf{Semantic} \\
\midrule
GS-Grouping       & 28.9 & 36.1 \\
LangSplat         & 13.6 & 41.2 \\
ReferSplat        & 37.4 & 24.4 \\
\rowcolor{rowblue}
\textbf{TrackRef3D (Ours)} & \textbf{48.5} & \textbf{68.3} \\
\bottomrule
\end{tabular}
\end{table}

\begin{table}[t]
\centering
\caption{Comparison on the LERF-OVS dataset with state-of-the-art methods in terms of \textbf{mIoU ($\uparrow$)}.}
\label{tab:lerf-ovs}
\setlength{\tabcolsep}{4.5pt}
\renewcommand{\arraystretch}{1.00}
\begin{tabular}{ l|
S[table-format=2.1,detect-weight=true,detect-inline-weight=math]
S[table-format=2.1,detect-weight=true,detect-inline-weight=math]
S[table-format=2.1,detect-weight=true,detect-inline-weight=math]
S[table-format=2.1,detect-weight=true,detect-inline-weight=math]
|
S[table-format=2.1,detect-weight=true,detect-inline-weight=math] }
\toprule
\textbf{Method} & \textbf{ram.} & \textbf{fig.} & \textbf{tea.} & \textbf{kit.} & \textbf{avg.} \\
\midrule
Feature-3DGS       & 43.7 & 58.8 & 40.5 & 39.6 & 45.7 \\
GS-Grouping     & 45.5 & 60.9 & 40.0 & 38.7 & 46.3 \\
LEGaussians    & 46.0 & 60.3 & 40.8 & 39.4 & 46.6 \\
GOI   & 52.6 & 63.7 & 44.5 & 41.4 & 50.6 \\
LangSplat   & 51.2 & 65.1 & 44.7 & 44.5 & 51.4 \\
ReferSplat          & 55.1 & 67.5 & 50.1 & 48.9 & 55.4 \\
\rowcolor{rowblue}
\textbf{TrackRef3D (Ours)}  & \textbf{63.0} & \textbf{74.3} & \textbf{59.1} & \textbf{56.3} & \textbf{63.2} \\
\bottomrule
\end{tabular}
\end{table}

\begin{table}[t]
\centering
\caption{Comparison on the 3D-OVS dataset with state-of-the-art methods in terms of \textbf{mIoU ($\uparrow$)}.}
\label{tab:3d-ovs}
\setlength{\tabcolsep}{2.7pt}
\renewcommand{\arraystretch}{1.00}
\begin{tabular}{ l|
S[table-format=2.1,detect-weight=true,detect-inline-weight=math]
S[table-format=2.1,detect-weight=true,detect-inline-weight=math]
S[table-format=2.1,detect-weight=true,detect-inline-weight=math]
S[table-format=2.1,detect-weight=true,detect-inline-weight=math]
S[table-format=2.1,detect-weight=true,detect-inline-weight=math]
|
S[table-format=2.1,detect-weight=true,detect-inline-weight=math] }
\toprule
\textbf{Method} & \textbf{Bed} & \textbf{Bench} & \textbf{Room} & \textbf{Sofa} & \textbf{Lawn} & \textbf{avg.} \\
\midrule
GS-Grouping       & 83.0 & 91.5 & 59.9 & 87.3 & 90.6 & 87.7 \\
Feature-3DGS     & 83.5 & 90.7 & 84.7 & 86.9 & 93.4 & 87.8 \\
GOI    & 89.4 & 92.8 & 91.3 & 85.6 & 84.1 & 90.6 \\
LEGaussians    & 84.9 & 91.1 & 86.0 & 87.8 & 92.5 & 88.5 \\
LangSplat            & 92.5  & 94.2 & 94.1 & 90.0 & 96.1 & 93.4 \\
ReferSplat           & 93.2 & 94.8 & 94.6 & 91.8 & 96.5 & 94.1 \\
\rowcolor{rowblue}
\textbf{TrackRef3D}  & \textbf{94.1} & \textbf{95.3} & \textbf{95.5} & \textbf{93.3} & \textbf{97.7} & \textbf{95.2} \\
\bottomrule
\end{tabular}
\end{table}

\begin{table}[t]
\centering
\caption{Ablation study on the key components.}
\label{tab:ablation1}
\setlength{\tabcolsep}{8.1pt}
\renewcommand{\arraystretch}{1.00}
\begin{tabular}{lcccc}
\toprule
\multirow{2.5}{*}{\textbf{Index}} & \multicolumn{2}{c}{\textbf{Components}} & \multicolumn{2}{c}{\textbf{Results}} \\  \cmidrule(l){2-3}\cmidrule(l){4-5}
& \textbf{TSCM} & \textbf{HTS} & \textbf{Ramen} & \textbf{Kitchen} \\
\midrule
Baseline  & \xmark & \xmark & 35.2 & 24.4 \\
1         & \cmark & \xmark & 39.4 & 29.1 \\
2         & \xmark & \cmark & 37.7 & 26.3 \\
\rowcolor{rowblue}
\textbf{Ours}      & \cmark & \cmark & \textbf{45.7} & \textbf{33.6} \\
\bottomrule
\end{tabular}
\end{table}

\begin{figure}[t]
  \centering
  \includegraphics[width=1.0\linewidth]{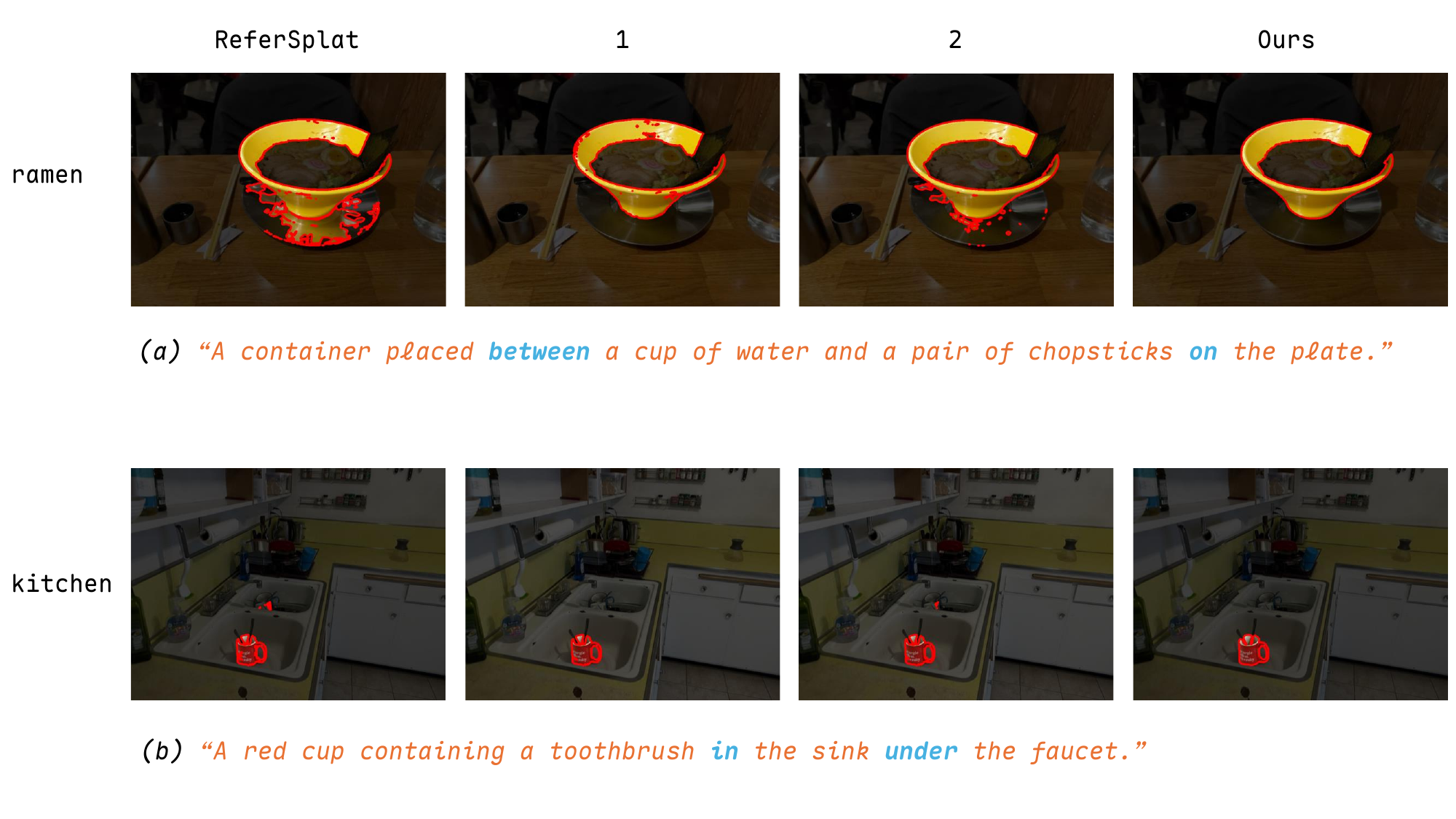}\vspace{0mm}
  \caption{Visualizations on the Ref-LERF dataset of the key components. The \textcolor[HTML]{46B1E1}{\textbf{blue}} highlights spatial cues.}
  \label{fig:ablation}
  \vspace{0mm}
\end{figure}

\begin{table}[t]
\centering
\setlength{\tabcolsep}{0.9pt}
\renewcommand{\arraystretch}{1.00}
\caption{Decomposed ablation study of TSCM on Ref-LERF.}
\label{tab:ablation2}
\begin{tabular}{lcc}
\toprule
\textbf{Method} & \textbf{Ramen} & \textbf{Kitchen} \\
\midrule
Baseline    & 35.2 & 24.4 \\
\midrule
(A): Florence-2 + SAM-2 & 23.1   & 13.7   \\
(B): (A) + tracking & 23.1   & 13.7   \\
(C): (B) + clustering only  &  31.7  & 22.8   \\
(D): (B) + voting only   & 39.3   & 27.1   \\
\rowcolor{rowblue}
\textbf{(E): (B) + voting + clustering (TSCM)}  & \textbf{45.7}   & \textbf{33.6}   \\
\bottomrule
\end{tabular}
\end{table}

\begin{table}[t]
\centering
\setlength{\tabcolsep}{20.2pt}
\renewcommand{\arraystretch}{1.00}
\caption{Ablation study on synonymous clustering threshold $\tau_{\text{sem}}$.}
\label{tab:ablation3}
\begin{tabular}{lcc}
\toprule
\textbf{Threshold} & \textbf{Ramen} & \textbf{Kitchen} \\
\midrule
0.70 & 29.6 & 20.4   \\
0.75 & 33.1 & 24.7 \\
0.80 & 39.5 & 28.4   \\
\rowcolor{rowblue}
\textbf{0.85} & \textbf{45.7}   & \textbf{33.6}   \\
0.90  &  41.3  & 28.5   \\
\bottomrule
\end{tabular}
\end{table}

\begin{table}[t]
\centering
\setlength{\tabcolsep}{13.0pt}
\renewcommand{\arraystretch}{1.00}
\caption{Ablation study on Description Generation.}
\label{tab:ablation4}
\begin{tabular}{lcc}
\toprule
\textbf{Method} & \textbf{Ramen} & \textbf{Kitchen} \\
\midrule
Maximum   & 19.4 & 12.0 \\
Minimum & 23.1   & 16.3   \\
Random   & 35.9   & 26.4   \\
Medium   & 40.3   & 29.2   \\
\midrule
Weighting ($\sigma = 50$)  & 39.8   & 29.4 \\
Weighting ($\sigma = 75$)  & 42.5   & 30.9 \\
\rowcolor{rowblue}
\textbf{Weighting ($\sigma = 100$)}  & \textbf{45.7}   & \textbf{33.6} \\
Weighting ($\sigma = 125$)  & 43.6   & 31.3 \\
\bottomrule
\end{tabular}
\end{table}

\textbf{Evaluation Metrics.} We adopt mean Intersection over Union (mIoU) as the primary evaluation metric due to the lack of 3D ground-truth annotations. Evaluation is performed by rendering the predicted 3D segmentation to 2D views and computing mIoU against 2D ground-truth masks. Specifically, we render segmentation masks from all unseen viewpoints and average the IoU scores for each query. For Ref-LERF and Laboratory scene, we evaluate on both short prompts and referring descriptions to assess robustness to varying query specificities. For short queries, we evaluate category-level masks from the union of all instances belonging to the queried category. For LERF-OVS and 3D-OVS, we follow the original open-vocabulary evaluation with category names as queries.

\begin{figure*}[t]
  \centering
  \includegraphics[width=1.0\linewidth]{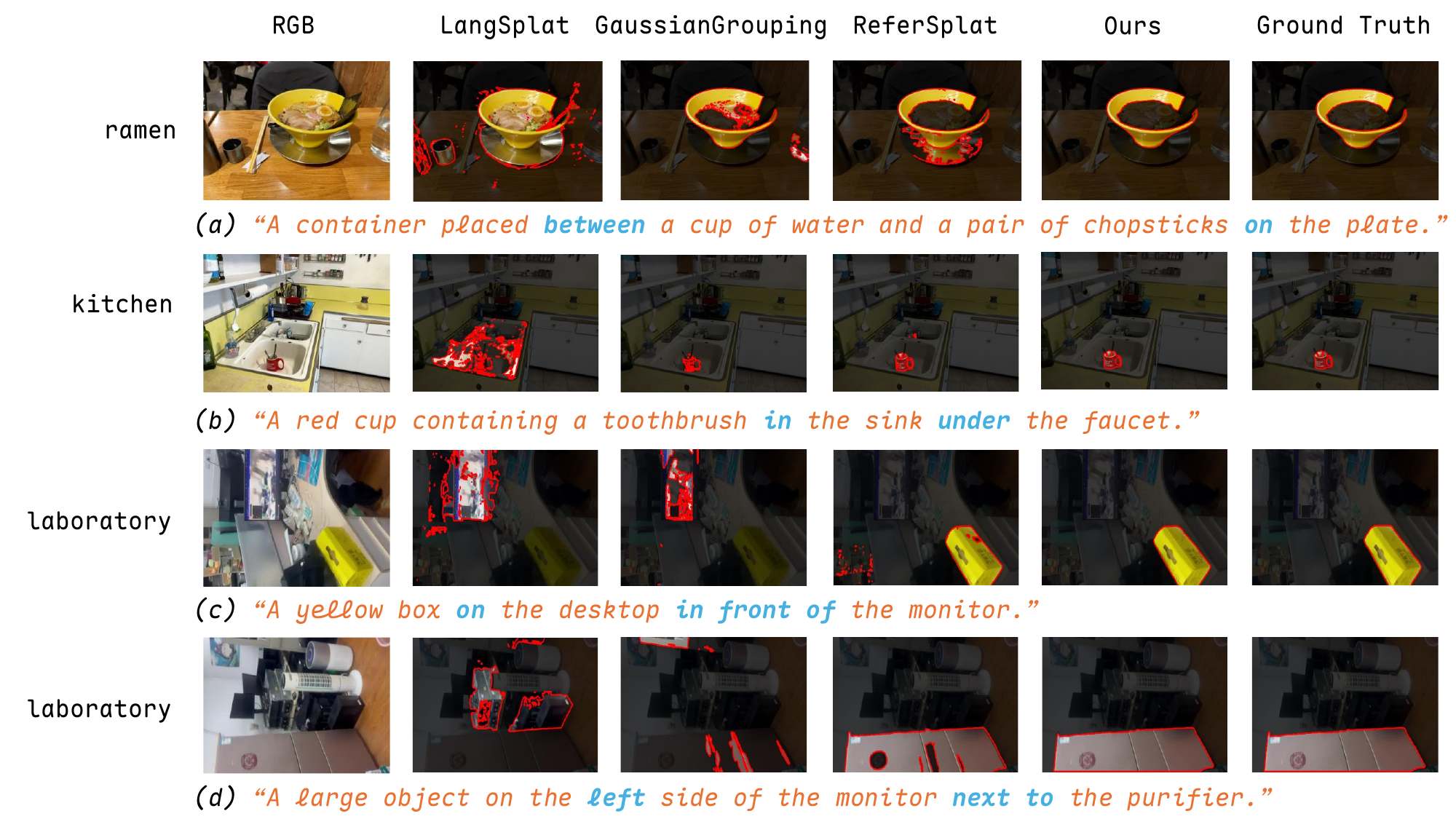}\vspace{0mm}
  \caption{Visualizations on the Ref-LERF dataset and our self-collected Laboratory scene. The \textcolor[HTML]{46B1E1}{\textbf{blue}} highlights spatial cues.}
  \label{fig:visualization}
  \vspace{0mm}
\end{figure*}

\textbf{Implementation Details.} Our method is trained on a single NVIDIA RTX 5880 Ada GPU with 48GB memory. We first use the vanilla 3DGS to reconstruct scene geometry from multi-view images, then build upon the ReferSplat codebase and extend it to open-world referring segmentation. We employ Florence-2 \cite{xiao2023florence} for open-vocabulary detection and SAM-2 \cite{ravi2024sam2} for segmentation to obtain per-view masks. DEVA \cite{cheng2023tracking} establishes cross-view object trajectories followed by synonymous clustering with threshold $\tau_{\text{sem}} = 0.85$ based on cosine similarity. We compute visibility scores using a Gaussian weighting with tolerance $\sigma = 100$ pixels to select the keyframe that balances object visibility and spatial context, then employ CogVLM-2 \cite{hong2024cogvlm2} to generate multi-granularity descriptions for each track. We combine binary cross-entropy segmentation loss with multi-positive contrastive loss ($\lambda = 0.1$, $\tau = 0.1$). Each object's training data includes short category semantics and multi-granularity descriptions embedded by BERT \cite{devlin2019bert}, all treated as positive pairs to enable coarse-to-fine learning. We train for 5 epochs using Adam optimizer \cite{kingma2015adam} with learning rates: position lr = $1.6 \times 10^{-4}$ decaying to $1.6 \times 10^{-6}$, feature lr = $2.5 \times 10^{-3}$, and MLP lr = $1 \times 10^{-4}$. During training, we apply a ratio decay schedule starting at 0.1 and multiplying by 0.6 every 2000 iterations.

\subsection{Results on Ref-LERF and Laboratory Scene}

\textbf{Quantitative Comparison.} Tab.~\ref{tab:ref-lerf} presents quantitative results on Ref-LERF benchmark, where we compare TrackRef3D with state-of-the-art methods on referring segmentation with natural language queries. Our method achieves substantial improvements across all four scenes, with 45.7 mIoU on Ramen and 33.6 mIoU on Kitchen, outperforming ReferSplat by 10.5 and 9.2 points, respectively. On average, TrackRef3D achieves 38.8 mIoU, surpassing ReferSplat by 9.6 points. This significant improvement demonstrates that our automatic pipeline can provide more consistent training signals than per-view pseudo supervision derived from manual annotations. Traditional 2D method Grounded SAM \cite{ren2024groundedsam} struggles with 3D consistency, while the neural field approach SPIn-NeRF \cite{spinnerf} achieves only 9.8 mIoU. Even recent 3D methods GS-Grouping \cite{gaussian_grouping}, GOI \cite{qu2024goi} and LangSplat \cite{qin2024langsplat} lag significantly behind. Tab.~\ref{tab:laboratory} presents results on our self-collected Laboratory scene. We evaluate on both short queries and referring descriptions to assess robustness across varying query specificities. For fair comparison, ReferSplat is trained using the same automatically generated supervision from our pipeline, and TrackRef3D differs from ReferSplat only by adopting the HTS. TrackRef3D achieves 48.5 mIoU on multi-granularity descriptions, outperforming ReferSplat by 11.1 points and demonstrating strong generalization to unseen environments. Critically, ReferSplat attains only 24.4 mIoU on short queries, as its training objective is dominated by referring descriptions and provides weak supervision for short queries, leading to poor short-query grounding. This validates the effectiveness that trained solely on referring descriptions suffers degradation on short queries.

\textbf{Qualitative Comparison.} Fig.~\ref{fig:visualization} presents visualizations on the Ref-LERF dataset and Laboratory scene. For referential queries, TrackRef3D accurately segments the target object with precise boundaries, while ReferSplat exhibits noisy artifacts and incomplete coverage due to multi-view inconsistency. TrackRef3D maintains strong segmentation quality, accurately identifying and segmenting the target objects. The visualizations strongly support our claim that trajectory-aware supervision improves cross-view consistency and reduces noisy artifacts.

\subsection{Results on the LERF-OVS and 3D-OVS Datasets}

We evaluate TrackRef3D on the LERF-OVS and 3D-OVS benchmarks. Tab.~\ref{tab:lerf-ovs} shows results on LERF-OVS, where TrackRef3D achieves 63.2 average mIoU. Tab.~\ref{tab:3d-ovs} presents results on the 3D-OVS benchmark, where TrackRef3D achieves 95.2 average mIoU. The results demonstrate the effectiveness of HTS in maintaining robust performance across varying query specificities.

\subsection{Ablation Study}

\textbf{Module Effectiveness.} We conduct an ablation study on Ref-LERF. As shown in Tab.~\ref{tab:ablation1} and Fig.~\ref{fig:ablation}, we evaluate on the ramen and kitchen scenes by progressively adding the TSCM and the HTS to the ReferSplat baseline. Without TSCM, the supervision is taken from ReferSplat's manual annotation and HTS is applied using the corresponding manual category labels and descriptions.

\textbf{Decomposed Ablation of TSCM.} To isolate the effectiveness of the TSCM, we conduct a controlled ablation where all variants from A to E are trained with the same descriptions generated from the same keyframe and optimized with the HTS, and  the Baseline row reports the original ReferSplat result. Tab.~\ref{tab:ablation2} presents a decomposed ablation of the TSCM on Ref-LERF. Replacing the per-view mask generation from the ReferSplat baseline (Grounded-SAM) with Florence-2 and SAM-2 in (A) yields lower performance compared to the ReferSplat baseline, suggesting that naively using per-view open-vocabulary predictions is insufficient for this benchmark due to cross-view inconsistency. Video tracking using DEVA alone in (B) does not improve the mIoU because it only establishes trajectory correspondences while keeping the per-view masks and labels unchanged. While both components of TSCM contribute to the final performance, the trajectory-aware voting in (D) yields a more substantial improvement than synonymous clustering in (C). The two components produce consistent supervision that significantly surpasses the ReferSplat baseline.

\textbf{Sensitivity to Threshold.} Tab.~\ref{tab:ablation3} evaluates different values of the synonymous clustering threshold $\tau_{\text{sem}}$. The small threshold tends to over-merge semantically related but distinct labels, which can blur category assignments and harm supervision quality. When the threshold becomes too large, clustering becomes conservative and misses beneficial synonymous clustering.

\textbf{Description Generation.} Tab.~\ref{tab:ablation4} ablates strategies for selecting the keyframe used to generate descriptions. Selecting the keyframe with extreme visibility performs poorly, confirming that too large or too small mask area is suboptimal for generating descriptions. Choosing the medium keyframe improves performance, while random selection yields intermediate results. We further adopt a Gaussian-like weighting over visibility scores to the keyframe, which outperforms heuristic baselines, with the best performance at $\sigma{=}100$.

\section{Conclusion}

We present TrackRef3D, a fully automatic pipeline for open-world referring segmentation in 3DGS without manual annotation. By introducing the multi-view consistent track-then-label paradigm, our approach fundamentally addresses multi-view inconsistency through the TSCM. Furthermore, the HTS ensures robustness under varying query specificities, bridging the gap between coarse category semantics and fine-grained referential cues. Extensive experiments demonstrate that TrackRef3D outperforms state-of-the-art methods, providing an open-world solution for referring segmentation in real-world applications.

\section*{Impact Statement}

This paper presents work whose goal is to advance the field of Machine
Learning. There are many potential societal consequences of our work, none of which we feel must be specifically highlighted here.

\bibliography{main}
\bibliographystyle{icml2026}
\end{document}